\documentclass[conference]{IEEEtran}
\IEEEoverridecommandlockouts
\usepackage{cite}
\usepackage{amsmath,amssymb,amsfonts}
\usepackage{algorithmic}
\usepackage{graphicx}
\usepackage{textcomp}
\usepackage{multirow}
\usepackage{booktabs}
\usepackage{xcolor}
\def\BibTeX{{\rm B\kern-.05em{\sc i\kern-.025em b}\kern-.08em
    T\kern-.1667em\lower.7ex\hbox{E}\kern-.125emX}}
\begin{document}

\title{Point Cloud Segmentation Using Transfer Learning with RandLA-Net: A Case Study on Urban Areas\\

}

\author{\IEEEauthorblockN{Alperen Enes Bayar\textsuperscript{1,3}, Ufuk Uyan\textsuperscript{1}, Elif Toprak\textsuperscript{1, 4},Cao Yuheng\textsuperscript{2}, Tang Juncheng\textsuperscript{2}, Ahmet Alp Kındıroğlu\textsuperscript{1}}
\IEEEauthorblockA{
\textit{\textsuperscript{1} Huawei Turkey R\&D Center} \\
\textit{\textsuperscript{2} Huawei Technologies} \\
\textit{\textsuperscript{3} Eskisehir Technical University } \\
\textit{\textsuperscript{4}Yeditepe University   } \\
\{alperen.enes.bayar1, ufuk.uyan2, elif.toprak, ahmet.alp.kindiroglu, tangjuncheng1, caoyuheng1\}@huawei.com,}
aebayar@ogr.eskisehir.edu.tr, elif.gumuslu@std.yeditepe.edu.tr

}

\maketitle

\begin{abstract}
Urban environments are characterized by complex structures and diverse features, making accurate segmentation of point cloud data a challenging task. 
This paper presents a comprehensive study on the application of RandLA-Net, a state-of-the-art neural network architecture, for the 3D segmentation of large-scale point cloud data in urban areas. 
The study focuses on three major Chinese cities, namely Chengdu, Jiaoda, and Shenzhen, leveraging their unique characteristics to enhance segmentation performance.

To address the limited availability of labeled data for these specific urban areas, we employed transfer learning techniques. 
We transferred the learned weights from the Sensat Urban and Toronto 3D datasets to initialize our RandLA-Net model. 
Additionally, we performed class remapping to adapt the model to the target urban areas, ensuring accurate segmentation results.

The experimental results demonstrate the effectiveness of the proposed approach  achieving over 80\% F1 score for each areas in 3D point cloud segmentation. 
The transfer learning strategy proves to be crucial in overcoming data scarcity issues, providing a robust solution for urban point cloud analysis. 
The findings contribute to the advancement of point cloud segmentation methods, especially in the context of rapidly evolving Chinese urban areas.
\end{abstract}


\section{Introduction}
Point cloud semantic segmentation plays a pivotal role in advancing structural representation learning and stereoscopic scene understanding within the realm of computer vision. 
This task involves partitioning the spatial context of a scene into semantically meaningful regions based on the inherent conformation and geometric knowledge encapsulated in point cloud layouts.
Notably, the success of point cloud semantic segmentation has found applications in critical domains such as autonomous driving, robotic manipulation, and virtual reality, prompting researchers to continually refine and enhance the accuracy of segmentation solutions.

Point clouds, capturing intricate geometric characteristics and surface context, serve as an indispensable representation for various 3D vision applications, including scene understanding, autonomous vehicles, and robotics. 
Point cloud segmentation, a key component of point cloud analysis, involves identifying points belonging to semantic categories of interest. 
However, the efficacy of segmentation models heavily relies on training data with point-level annotations, leading to high annotation costs. 
To address this challenge, weakly supervised methods have been developed, leveraging different weak supervisory signals such as partially labeled points, subcloud level annotations, or scene level annotations.

Additionally, it's worth noting that point cloud registration stands as a fundamental technique for LiDAR-based vehicle localization in urban road scenes, with established categorizations into local and global search methods within the SLAM (Simultaneous Localization and Mapping) community.
Our work seeks to contribute modestly to the ongoing advancements in segmentation. 
Moreover, it aims to align with the broader landscape of point cloud applications, with a specific emphasis on highlighting the potential relevance of inter-cloud semantics in the context of weakly supervised segmentation.

In this regard, our proposed methodology modestly attempts to extend the current boundaries of research. 
It offers a holistic approach intended to address challenges not only in semantic segmentation but also in the related field of point cloud registration.
By exploring the incorporation of inter-cloud semantics, we hope to contribute to the ongoing dialogue surrounding these complex issues in a meaningful way.
\section{Literature Review}

\begin{figure}[!t]
\includegraphics[width=8cm]{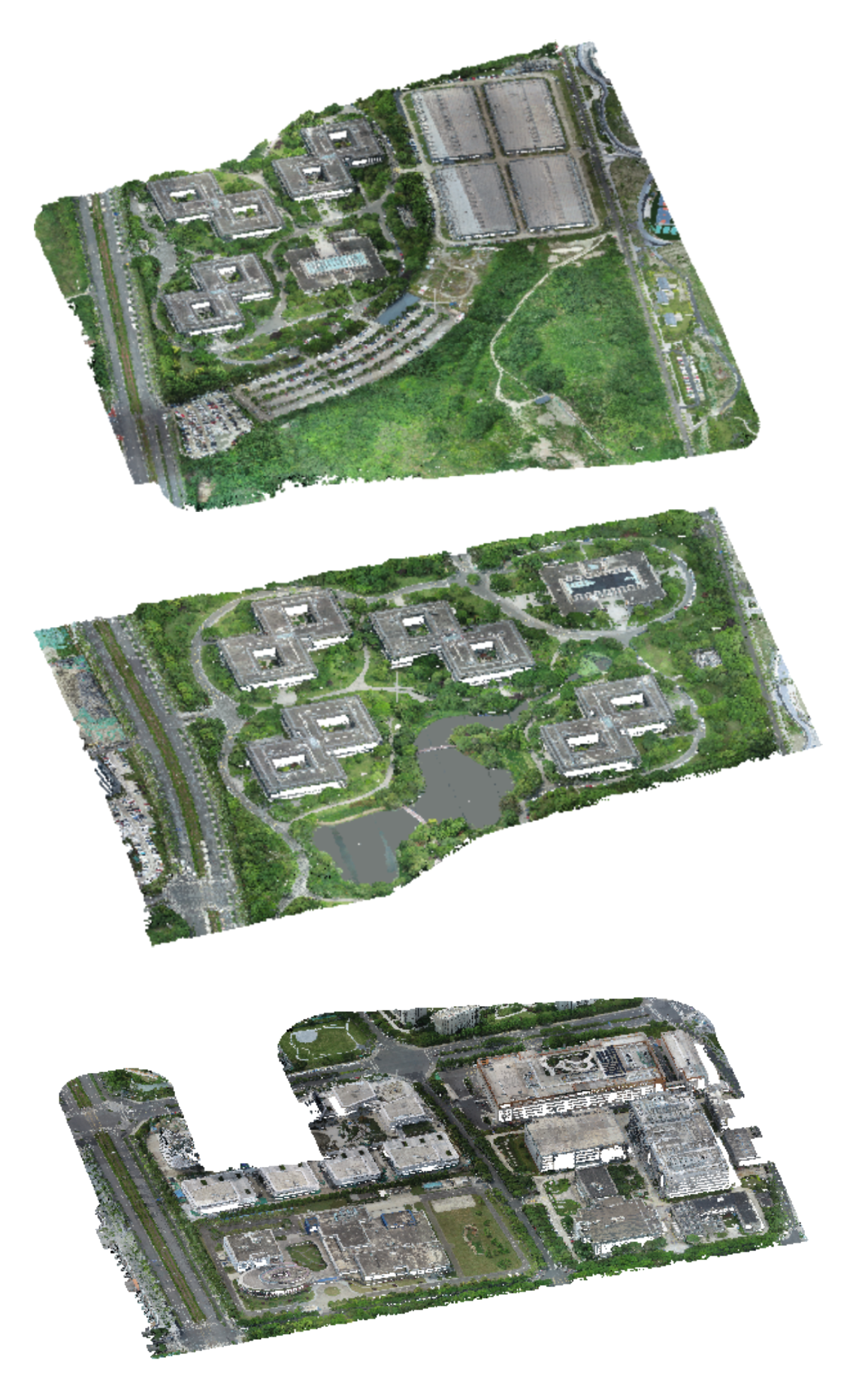}
\caption{Example images of Chengdu Point Cloud Dataset}
\label{fig:originals}
\end{figure}

Semantic segmentation of LiDAR point clouds is a critical task, particularly in the context of autonomous driving, where the demand for accurate and label-efficient approaches is paramount. 
Several recent works have made notable contributions to this field, addressing challenges associated with dataset labeling, weak supervision, and the ability to adapt to real-world scenarios.

Liu et al. proposed a label-efficient semantic segmentation pipeline tailored for outdoor scenes with LiDAR point clouds~\cite{liu2022less}. 
Acknowledging the challenges posed by the cost of labeling large datasets, the authors presented an approach that combines efficient labeling processes with semi/weakly supervised learning. The method leverages geometry patterns in outdoor scenes for heuristic pre-segmentation, reducing manual labeling requirements. 

The use of prototype learning and multi-scan distillation enhances point embeddings, exploiting richer semantics from temporally aggregated point clouds.
Remarkably, even with extremely limited human annotations (0.1\% point labels), the proposed method outperforms existing label-efficient techniques, showcasing competitiveness with fully supervised counterparts.

\begin{figure*}[h]
    \includegraphics[width=\linewidth]{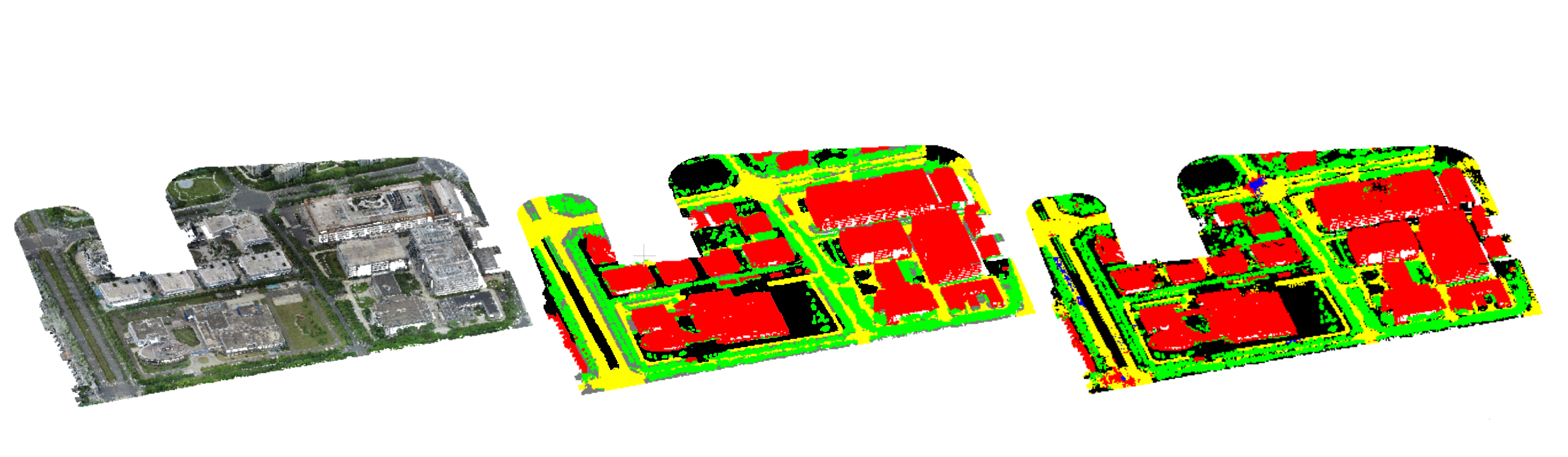}
    \caption{From left to right; original, ground truth and prediction of Chendu Area 3}
    \label{fig:comp}
\end{figure*}

In the realm of weakly supervised 3D semantic instance segmentation, Chibane et al. introduced Box2Mask, a method that utilizes 3D bounding box labels as an alternative to dense per-point annotations~\cite{chibane2023box2mask}. 
The authors proposed a deep model inspired by classical Hough voting, directly voting for bounding box parameters. A clustering method, tailored to bounding box votes, goes beyond center votes, demonstrating superior performance among weakly supervised approaches on the ScanNet test. 
The practicality of the approach is highlighted by its successful training on the ARKitScenes dataset, annotated with 3D bounding boxes only, showcasing compelling 3D instance segmentation masks.

Addressing the limitations of current LIDAR semantic segmentation methods, Cen et al. proposed the open-world semantic segmentation task, targeting robustness and adaptability for real-world applications such as autonomous driving~\cite{cen2022openworld}. 
The closed-set and static nature of current methods restrict their applicability to novel objects and evolving scenarios. 
The REdundAncy cLassifier (REAL) framework was introduced, offering a general architecture for both open-set semantic segmentation and incremental learning. 
Experimental results demonstrate state-of-the-art performance in open-set semantic segmentation on SemanticKITTI and nuScenes datasets, while also effectively mitigating the catastrophic forgetting problem during incremental learning.

\section{Method}
\subsection{Datasets}
\subsubsection{Chengdu Point Cloud Segmentation Dataset}
As one of the main contributions of this paper, we have collected and manually annotated a large scale pointcloud segmentation dataset. The dataset consists of point cloud data and annotations collected from Chengdu, Jiaoda, and Shenzhen regions, as can be seen in Table~\ref{tab:points}.

To enhance the granularity and accuracy of the dataset, we have subdivided these areas into smaller chunks, facilitating a more detailed analysis. 
In the preprocessing phase, we employed the Open3D library to remove noise and outliers through statistical and radius-based techniques, ensuring the quality of the point cloud data. Figure~\ref{fig:originals} shows an example area in the Chengdu Point Cloud Segmentation Dataset.

The segmentation task involves classifying points into five distinct categories: Background, Building, Vegetation, Road, and Water. 
The richness of this dataset extends beyond geographical diversity; it captures the contextual intricacies of urban landscapes. 
This characteristic presents a challenging yet realistic scenario for developing and evaluating segmentation models.

In the preprocessing stage, we manually refined the water areas to enhance the accuracy of annotations. 
This meticulous editing process ensures the dataset's reliability for training and evaluating models, particularly in regions where automatic algorithms may face challenges. 
The Chengdu Point Cloud Segmentation Datasets, with their detailed annotations and carefully curated preprocessing steps, contribute significantly to the robustness and accuracy of our segmentation model. 

\begin{table}[b]
    \centering
    \caption{Number of Points of each Area in Chengdu Point Cloud Dataset}
    \begin{tabular}{p{0.8cm}p{0.8cm}|p{1cm}p{0.9cm}p{0.9cm}p{0.9cm}p{0.9cm}}
        \toprule
        Location & Subarea & Background & Building & Vegetation & Road     & Water   \\
        \midrule
        \multirow{3}{*}{Chengdu}   & Area 1  & 35890436   & 15779239 & 40057320   & 16273005 & 0       \\
                                   & Area 2  & 22398111   & 9156839  & 19363119   & 12821506 & 4260425 \\
                                   & Area 3  & 13568733   & 13640479 & 5786533    & 9004255  & 0       \\
        \multirow{3}{*}{Jiaoda}    & Area 1  & 11953153   & 6223082  & 14104466   & 8031482  & 687817  \\
                                   & Area 2  & 6870420    & 2693957  & 5030479    & 1433871  & 971273  \\
                                   & Area 3  & 12197832   & 9654062  & 17648096   & 4543560  & 956450  \\
        Shenzhen           & Area 1 & 17864435   & 12283776 & 32280982   & 5570807  & 0       \\
        \bottomrule
    \end{tabular}
\label{tab:points}
\end{table}

\subsubsection{Public Datasets}
In addition to the proprietary Chengdu Point Cloud Segmentation Datasets, we leveraged publicly available datasets to enhance the generalizability and robustness of our segmentation model. 
The public datasets encompass a broader scope of urban environments, allowing for a more comprehensive understanding of point cloud segmentation challenges. 
Notably, these datasets may have a more extensive set of classes, necessitating a class remapping process to align them with our specific segmentation requirements.
The remapping involves converting the varied classes present in public datasets to the standardized five classes used in our research: Background, Building, Vegetation, Road, and Water.

SensatUrban~\cite{hu2021semantic} is a notable urban-scale photogrammetric point cloud dataset, comprising nearly three billion richly annotated points. 
This dataset represents a significant advancement, providing five times the number of labeled points compared to existing large point cloud datasets. 
Covering extensive areas from two UK cities, SensatUrban spans approximately 6 km2 of the city landscape. 
Each 3D point in the dataset is meticulously labeled with one of 13 semantic classes, including ground, vegetation, car, and more.

Toronto-3D~\cite{Tan_2020} is a large-scale urban outdoor point cloud dataset acquired using a Mobile Laser Scanning (MLS) system in Toronto, Canada, specifically designed for semantic segmentation tasks. 
The dataset covers approximately 1 km of road and consists of around 78.3 million points. 
It serves as a valuable resource for understanding semantic segmentation challenges in urban environments, providing detailed annotations for various classes.

This dual-source approach to dataset utilization ensures that our segmentation model is trained and evaluated on both proprietary and publicly accessible data, providing a more comprehensive assessment of its performance across different urban contexts. 
The remapping process facilitates the integration of valuable insights from diverse datasets into a unified framework, contributing to the development of a robust and adaptable point cloud segmentation model tailored for Chinese urban areas.

\subsection{Preprocessing}

In preparation for effective point cloud segmentation, a preprocessing pipeline was employed. 
This pipeline comprises several key steps aimed at ensuring uniformity, compatibility, and accuracy in the subsequent segmentation model.

\textbf{Converting Batches to Same Size:} To create a consistent and manageable dataset, each batch was standardized to the same size. 
We generated an equal number of points for each batch, set to a predefined value, such as 1e6. 
This uniformity facilitates a streamlined training process and ensures that each batch is of a consistent size, enabling efficient model convergence. 
Additionally, the entire dataset was subdivided into smaller chunks, facilitating more manageable and parallelized processing. 
As part of the RandLA-Net model preprocessing, we converted the data format, aligning it with the model's requirements.

\textbf{Class Remapping:} Publicly available datasets often come with diverse class labels, necessitating a class remapping process to align them with our standardized five classes: Background, Building, Vegetation, Road, and Water. 
For instance, in the SensatUrban dataset, the original class labels were remapped to our predefined classes, such as Ground, Vegetation, Building, Wall, Bridge, Parking, Rail, Traffic Road, Street Furniture, Car, Footpath, Bike, and Water.
Similarly, for the Toronto3D dataset, class labels were remapped to our classes, including Unclassified, Ground, Road Markings, Natural, Building, Utility Line, Pole, Car, and Fence. 
This remapping ensures consistency in class representation across different datasets, contributing to a unified segmentation model.

\textbf{Number of Neighbours:} A critical aspect of point cloud preprocessing involves exporting a graph-like tree structure that contains neighborhood information. 
This tree structure encapsulates the relationships and spatial dependencies among points, providing essential contextual information for the segmentation model. 
The conversion from PLY to NPY format ensures compatibility with the RandLA-Net model's processing requirements, enhancing the model's ability to understand and leverage the spatial relationships within the point cloud data during the segmentation process.

These preprocessing steps collectively contribute to the efficiency, accuracy, and generalization capabilities of our point cloud segmentation model, preparing the data for effective training and evaluation on diverse urban environments.

\subsection{3D Segmentation}

RandLA-Net~\cite{hu2020randlanet}, which stands for Randomized Aggregating of Local Features for Large-Scale Point Cloud Semantic Segmentation, is a neural network architecture designed for efficient semantic segmentation of large-scale 3D point clouds. The primary challenge addressed by RandLA-Net is the ability to process massive point clouds efficiently without compromising on semantic segmentation accuracy.

\subsubsection{Random Sampling for Efficiency}
The core innovation of RandLA-Net lies in its use of random point sampling, as opposed to more complex and computationally expensive point selection methods. Random sampling allows the network to significantly decrease point density without sacrificing computational efficiency. However, random sampling introduces the risk of discarding important features by chance.

\subsubsection{Local Feature Aggregation}
To mitigate the potential loss of key features due to random sampling, RandLA-Net incorporates a novel Local Feature Aggregation module. This module comprises three neural units:

\subsubsection{Local Spatial Encoding (LocSE)}
Gathers neighboring points using the K-nearest neighbors (KNN) algorithm based on Euclidean distances.
Encodes relative point positions to make the network aware of local geometric patterns.
Augments point features with encoded relative point positions, enhancing the overall feature vector.

\subsubsection{Attentive Pooling}
Utilizes an attention mechanism to automatically learn important local features.
Computes attention scores for each feature using a shared function followed by softmax.
Performs weighted summation of features based on the learned attention scores.

\subsubsection{Dilated Residual Block}
Stacks multiple LocSE and Attentive Pooling units with a skip connection to create a dilated residual block.
Increases the receptive field for each point, allowing the preservation of geometric details even after substantial downsampling.
Balances the trade-off between efficiency and effectiveness by stacking two sets of LocSE and Attentive Pooling as the standard residual block.
\subsection{Postprocessing}
\begin{figure*}[h]
    \includegraphics[width=\linewidth]{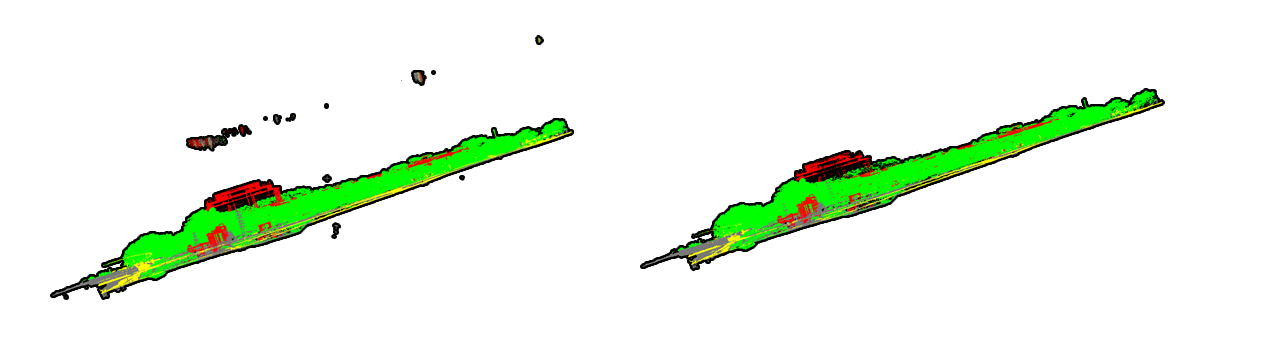}
    \caption{Before and After postprocessing applied it to the Point Clouds}
    \label{fig:post}
\end{figure*}

After the initial segmentation with RandLA-Net, postprocessing steps were employed to enhance the quality of the segmentation results and address specific challenges associated with 3D point clouds. 
In this section, we discuss two crucial postprocessing techniques: noise removal and height-based filtering. Figure~\ref{fig:post} shows an example of the postprocessing output.

\subsubsection{Noise Removal}
Noise in 3D point clouds can arise from various sources, including sensor inaccuracies, environmental factors, or artifacts during data acquisition. Noise can adversely affect the segmentation results by introducing spurious points or misclassifying legitimate points. Therefore, noise removal is a critical postprocessing step to improve the overall accuracy of the segmentation.

Several noise removal techniques can be applied to refine the segmented point cloud:

\textbf{Statistical Outlier Removal:} Utilizes statistical measures, such as mean and standard deviation, to identify and remove points deviating significantly from the expected distribution~\cite{ning2018efficient}.

\textbf{Voxel Grid Downsampling:} Divides the point cloud into voxels and retains a representative point from each voxel, effectively reducing the overall point density and removing redundant information~\cite{jung2020automated}.

\textbf{Radius Outlier Removal:} Eliminates points that have fewer neighbors within a specified radius, helping to filter out isolated noise points~\cite{bustos2017guaranteed}.

\textbf{Morphological Operations:} Applies morphological operations, such as erosion and dilation, to smooth the point cloud and eliminate small isolated clusters of noise~\cite{balado2020mathematical}.

\subsubsection{Height-Based Filtering}
Height-based filtering is particularly relevant in scenarios where the vertical dimension holds significant semantic information. 
For instance, in urban environments, the height of points above ground level may correspond to distinct objects or terrain features. 

\textbf{Height Thresholding:} Establishes height thresholds to filter out points that fall below or above certain elevations, aiding in the isolation of objects or terrain features of interest~\cite{rashidi2017ground}.

\textbf{Local Height Variation:} Analyzes local height variations to identify and preserve structures with significant elevation changes, contributing to the discrimination of distinct objects~\cite{he2015novel}.

\section{Experiments \& Results}
\subsection{Experimental Setup}
\subsubsection{Training Configuration}
The RandLA-Net model is trained using a dedicated workstation equipped with a high-performance GPU. The training process involves optimizing the model parameters using the Adam optimizer with a learning rate of 0.001. We employ batch normalization to enhance model stability and incorporate dropout layers to prevent overfitting. The training is performed for a specified number of epochs with early stopping criteria based on validation performance.

\subsubsection{Transfer Learning Strategy}
Given the limited availability of labeled data for the target Chinese urban areas (Chengdu, Jiaoda, Shenzhen), we adopt a transfer learning strategy. The RandLA-Net model is initialized with weights pre-trained on the Sensat Urban and Toronto 3D datasets. This transfer learning approach leverages knowledge gained from diverse urban contexts to bootstrap the segmentation model for the specific Chinese urban areas.

\subsubsection{Class Remapping}
To adapt the RandLA-Net model to the target urban areas, we performed class remapping. The original classes present in the proprietary Chengdu Point Cloud Segmentation Datasets were mapped to the standardized five classes used in our research: Background, Building, Vegetation, Road, and Water. This ensures consistency in class representation across different datasets and facilitates a unified evaluation framework.

\subsection{Evaluation Metrics}
The segmentation performance of RandLA-Net was assessed using standard metrics for point cloud segmentation:

\subsubsection{Intersection over Union (IoU)}
IoU measures the overlap between the predicted segmentation and ground truth. It is computed as the ratio of the intersection to the union of the predicted and ground truth regions.
\[ IoU = \frac{TP}{TP + FP + FN} \]
where \(TP\) is the true positive, \(FP\) is the false positive, and \(FN\) is the false negative.

\subsubsection{Accuracy}
Overall accuracy provides a holistic measure of the model's ability to correctly classify points across all classes.
\[ Accuracy = \frac{TP + TN}{TP + FP + TN + FN} \]
where \(TN\) is the true negative.

\subsubsection{F1 Score}
F1 score is the harmonic mean of precision and recall, providing a balanced measure of a model's performance.
\[ F1 = \frac{2 \cdot \text{Precision} \cdot \text{Recall}}{\text{Precision} + \text{Recall}} \]
where Precision is the ratio of true positive predictions to the total predicted positives, and Recall is the ratio of true positive predictions to the total actual positives.

\subsection{Experimental Results}
The experimental results showcase the effectiveness of the proposed approach for 3D point cloud segmentation in Chinese urban areas using the RandLA-Net model. The segmentation performance is evaluated across three major cities: Chengdu, Jiaoda, and Shenzhen. Figure~\ref{fig:comp} shows an example of the output of the RandLA-Net model.

In the evaluation metrics, key performance indicators such as Intersection over Union (IoU), accuracy, and F1 score are reported for different classes in each urban area. 
The results in Table~\ref{tab:results} indicate high accuracy and segmentation quality across diverse urban landscapes, affirming the model's capability to generalize to Chinese urban areas.

Notably, the transfer learning strategy proves crucial in overcoming limited labeled data availability, showcasing its effectiveness in enhancing segmentation performance. 
The class remapping process ensures consistency in the representation of semantic classes across different datasets, facilitating a unified evaluation framework.

The findings contribute to the advancement of point cloud segmentation methods, particularly in the context of rapidly evolving Chinese urban areas. 
The proposed approach addresses challenges associated with data scarcity and diverse urban characteristics, providing a reliable solution for accurate 3D point cloud analysis. 
The comprehensive study lays the groundwork for further research in semantic segmentation and urban mapping, fostering advancements in computer vision applications such as network planing, autonomous driving, robotic manipulation, and virtual reality within complex urban environments.

\begin{table*}[ht]
    \centering
    \caption{Results for each Area in Chengdu Point Cloud Dataset }
    \begin{tabular}{ll|ccc|ccc|ccc|ccc|ccc}
        	\toprule
        Location & Subarea & \multicolumn{3}{c}{Background} & \multicolumn{3}{c}{Building} & \multicolumn{3}{c}{Vegetation} & \multicolumn{3}{c}{Road} & \multicolumn{3}{c}{Water} \\
        \cmidrule(lr){3-5} \cmidrule(lr){6-8} \cmidrule(lr){9-11} \cmidrule(lr){12-14} \cmidrule(lr){15-17}
        && Acc & IoU & F1 & Acc & IoU & F1 & Acc & IoU & F1 & Acc & IoU & F1 & Acc & IoU & F1 \\
        \midrule
        \multirow{3}{*}{Chengdu} 
        & Area 1  & 0.98   & 0.90 & 0.95   & 0.99 & 0.94 & 0.97   & 0.97 & 0.96 & 0.98 & 0.84 & 0.81 & 0.89 & - & - & - \\
        & Area 2  & 0.98    & 0.94  & 0.97   & 0.98 & 0.92 & 0.96   & 0.98 & 0.97 & 0.99 & 0.90 & 0.85 & 0.92 & 0.92 & 0.87 & 0.93 \\
        & Area 3  & 0.97   & 0.91 & 0.95    & 0.94  & 0.91 & 0.95   & 0.99 & 0.93 & 0.97 & 0.88 & 0.83 & 0.91 & - & - & - \\
        \multirow{3}{*}{Jiaoda} 
        & Area 1  & 0.93   & 0.90  & 0.95   & 0.94  & 0.84 & 0.91   & 0.99 & 0.95 & 0.97 & 0.89 & 0.88 & 0.94 & 0.98 & 0.85 & 0.92 \\
        & Area 2  & 0.93    & 0.93  & 0.96    & 0.98  & 0.93 & 0.96   & 0.99 & 0.93 & 0.96 & 0.87 & 0.85 & 0.92 & 0.99 & 0.90 & 0.95 \\
        & Area 3  & 0.83   & 0.78  & 0.88   & 0.91  & 0.77 & 0.87   & 0.99 & 0.94 & 0.97 & 0.82 & 0.71 & 0.83 & 0.99 & 0.89 & 0.94 \\
        \multirow{1}{*}{Shenzhen} 
        & Area 1  & 0.92   & 0.80  & 0.89   & 0.77  & 0.70 & 0.82   & 0.98 & 0.93 & 0.96 & 0.68 & 0.59 & 0.75 & - & - & - \\
        \midrule
    \end{tabular}
    \label{tab:results}
\end{table*}

\section{Conclusion}
In conclusion, this paper has presented a comprehensive study on the application of RandLA-Net for the 3D segmentation of large-scale point cloud data in Chinese urban areas. 
Focusing on Chengdu, Jiaoda, and Shenzhen, the research leveraged the unique characteristics of these cities to enhance segmentation performance. 
The study addressed the challenge of limited labeled data by employing transfer learning techniques, initializing RandLA-Net with weights from Sensat Urban and Toronto 3D datasets, and performing class remapping for accurate segmentation results.

The experimental results demonstrate the effectiveness of the proposed approach in achieving high-quality 3D point cloud segmentation. 
The transfer learning strategy played a crucial role in overcoming data scarcity issues, providing a robust solution for urban point cloud analysis. The findings contribute significantly to the advancement of point cloud segmentation methods, particularly in the rapidly evolving Chinese metropolitan areas.

The paper contributes to the existing literature by showcasing the adaptability and robustness of RandLA-Net in the context of Chinese urban environments.
The dual-source utilization of proprietary Chengdu datasets and publicly available datasets ensures a comprehensive evaluation of the model's performance across different urban contexts. 
The presented preprocessing steps, including class remapping and dataset refinement, contribute to the efficiency and accuracy of the segmentation model.

In summary, this research provides valuable insights into the state-of-the-art techniques for 3D point cloud segmentation, with a specific focus on Chinese urban areas. 
The findings have practical implications for applications such as network planing, autonomous driving, robotic manipulation, and urban mapping. 
The presented methodology and experimental results contribute to the ongoing advancements in the field, opening avenues for further research in the intersection of point cloud analysis and urban dynamics.

\bibliographystyle{plain} 
\bibliography{refs} 

\end{document}